\documentclass[letterpaper, 10 pt, conference]{ieeeconf}  

\IEEEoverridecommandlockouts                              

\usepackage{graphicx} 
\usepackage{float}
\usepackage{amsmath} 
\usepackage{url}

\usepackage{booktabs}
\usepackage{balance}

\title{\LARGE \bf
Nano-U: Efficient Terrain Segmentation for Tiny Robot Navigation
}

\author{Federico Pizzolato, Francesco Pasti and Nicola Bellotto
\thanks{All authors are with Dept\,of\,Information\,Engineering, University\,of\,Padua, Italy. 
Contact author: N.~Bellotto ({\tt\small nbellotto@dei.unipd.it})}%
}

\begin{document}

\maketitle
\thispagestyle{empty}
\pagestyle{empty}

\begin{abstract}
Terrain segmentation is a fundamental capability for autonomous mobile robots operating in unstructured outdoor environments. However, state-of-the-art models are incompatible with the memory and compute constraints typical of microcontrollers, limiting scalable deployment in small robotics platforms. 
To address this gap, we develop a complete framework for robust binary terrain segmentation on a low-cost microcontroller.
At the core of our approach we design \mbox{Nano-U}, a highly compact binary segmentation network with a few thousand parameters.
To compensate for the network's minimal capacity, we train Nano-U via Quantization-Aware Distillation~(QAD), combining knowledge distillation and quantization-aware training.
This allows the final quantized model to achieve excellent results on the Botanic Garden dataset and to perform very well on TinyAgri, a custom agricultural field dataset with more challenging scenes.
We deploy the quantized \mbox{Nano-U} on a commodity microcontroller by extending MicroFlow, a compiler-based inference engine for TinyML implemented in Rust.
By eliminating interpreter overhead and dynamic memory allocation, the quantized model executes on an \mbox{ESP32-S3} with a minimal memory footprint and low latency.
This compiler-based execution demonstrates a viable and energy-efficient solution for perception on low-cost robotic platforms.

\end{abstract}

\section{INTRODUCTION}
\label{sec:intro}

Terrain segmentation is a fundamental perception capability for autonomous mobile robots operating in unstructured outdoor environments such as agricultural fields and natural trails~\cite{guan2022ganav,liu2023botanicgarden}.
Unlike structured road scenarios, traversability estimation in these settings is challenging due to dense vegetation, uneven ground, and strong lighting variation.
While deep learning has become the standard for robust segmentation, state-of-the-art models are impractical for tiny robotic platforms subject to strict kilobyte memory budgets and milliwatt power constraints~\cite{neuman2022tinyrobotlearning}.
Adapting these computationally heavy models to run on low-power microcontrollers~(MCUs) is a fundamental challenge for the scalable deployment of small and inexpensive mobile robots.

Existing Tiny Machine Learning~(TinyML) solutions typically target simpler classification tasks or rely on dedicated ML accelerators~\cite{abadade2023comprehensive}.
In this work instead, we use a commodity, general-purpose microcontroller board ESP32-S3-CAM, proposing a framework for robust terrain segmentation on non-specialized, widely accessible platforms costing as little as \$10.
To this end, our main contributions are as follows:

\begin{figure}
    \centering
    \includegraphics[width=0.49\textwidth]{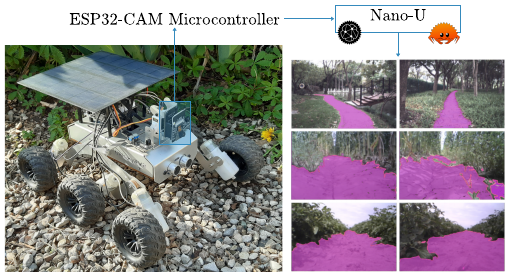}
    \caption{The Nano-U segmentation model identifies traversable terrains~(Right). It executes entirely on an ESP32-S3 microcontroller via the Rust-based MicroFlow engine, demonstrating feasibility for the tight memory and compute resources of a small mobile robot~(Left).}
    \label{fig:rover}
\end{figure}

\begin{itemize}
\item We design Nano-U, a small (33~KB with integer INT8 precision) binary segmentation network inspired by the popular U-Net architecture.
To fit within strict memory limits, we deliberately omit some standard connections and overcome accuracy losses by training Nano-U via adapting Quantization-Aware Distillation~(QAD)~\cite{polino2018dist_and_quant, xin2026nvidia_qad}. 
This single-pass regime fuses knowledge distillation from a full-scale teacher with INT8 fake-quantization, enabling the network to learn robust representations despite its extreme parameter constraints.

\item We deploy Nano-U by extending MicroFlow~\cite{carnelos2025microflow}, a compiler-based Rust inference engine for TinyML.
By resolving the network topology at compile time, we completely bypass interpreter overhead and dynamic memory allocations during inference. Furthermore, this bare-metal approach eliminates unnecessary peripheral initializations, maximizing available internal RAM and strictly bounding energy consumption.
\end{itemize}

We evaluate Nano-U on Botanic Garden~\cite{liu2023botanicgarden}, a comprehensive dataset for outdoor robot navigation, and on TinyAgri, a custom  and challenging terrain-segmentation dataset collected via the onboard camera of a small rover~(Fig.~\ref{fig:rover}).
We publicly release Nano-U and the TinyAgri dataset to support further research in edge robotics applications\footnote{{https://github.com/federico-pizz/Nano-U}}.
To validate our framework's efficiency, we also conduct hardware profiling directly on the ESP32-S3-CAM during Nano-U inference, revealing low RAM utilization, energy consumption, and processing latency.

The remainder of the paper is organized as follows:
Sec.~\ref{sec:related} reviews related work;
Sec.~\ref{sec:system} describes \mbox{Nano-U} and the optimization pipeline;
Sec.~\ref{sec:exp} details the experimental setup;
Sec.~\ref{sec:results} presents the results of the segmentation and hardware performance;
Sec.~\ref{sec:conclusions} summarizes the main achievements and suggests opportunities for future research.

\section{RELATED WORK}
\label{sec:related}

\subsection{Terrain Segmentation in Robotics}
\label{sec:segmentation}

Before the widespread adoption of deep learning~(DL), traversable terrain was commonly identified through hand-crafted appearance models, such as self-supervised color statistics propagated from a seed region ahead of the robot~\cite{dahlkamp2006selfsuper}.
Early DL approaches instead replaced the classification head of standard CNNs with convolutional layers to enable dense per-pixel prediction~\cite{long2015fully_cnn}. 
Subsequent encoder-decoder architectures such as SegNet~\cite{Badrinarayanan2017SegNet} and \mbox{U-Net}~\cite{ronneberger2015unet} recovered spatial detail lost during downsampling via skip connections. 
DeepLabV3 and its extension~\cite{chen2018deeplabv3plus,chen2017deeplab} instead captured multi-scale context through Atrous Spatial Pyramid Pooling~(ASPP), coupling it with a lightweight decoder to recover sharp object boundaries.

However, all the above models assume GPU-based inference, with sizes ranging from tens to hundreds of millions of parameters.
While these architectures have been successfully applied to terrain segmentation in agricultural robotics~\cite{milioto2018realtime_seg, lottes2018stem_detection}, their substantial resource requirements make them incompatible with the constraints of small robots~\cite{neuman2022tinyrobotlearning}. 
To satisfy the strict memory constraints of commodity MCUs, we depart from this multi-class paradigm and focus on binary traversability estimation, a simplified formulation that remains fully sufficient for autonomous path planning~\cite{guan2022ganav}.

\subsection{Efficient Neural Networks}
\label{sec:efficient}

Deploying neural networks on constrained hardware has driven research into architectures and training methods that reduce model size without significant accuracy loss.
At the architectural level, traditional works scale down model parameters and computational overhead by decreasing network depth and scaling channel width multipliers. 
This is often done via techniques such as Neural Architecture Search~\cite{salmani2025reviewonnas}, which automates the exploration of architecture configurations within specified hardware budgets.
Beyond it, optimizing the fundamental computational blocks is critical for on-device efficiency~\cite{howard2017mobilenets, sandler2018mobilenetv2}.

However, scaling down an architecture limits its representational capacity.
Knowledge distillation~\cite{hinton2015kd} mitigates this by transferring the generalization capability of a large teacher network into a compact student by matching soft output distributions rather than hard ground-truth labels.
Temperature scaling flattens the teacher's output, exposing gradient signal on ambiguous boundary pixels that a binary label would suppress.

Finally, neural networks are typically quantized to INT8 precision to be executed more efficiently on MCU-based embedded systems.
However, this introduces a significant accuracy degradation when high-precision floating-point FP32 weights are mapped to INT8 after training.
Quantization-Aware Training~(QAT)~\cite{jacob2018quantization_integer} addresses this by injecting fake-quantization nodes that simulate 8-bit rounding in the forward pass while preserving full-precision gradients via the straight-through estimator.
This is essential for small networks, where the representational margin is too narrow to absorb the perturbation of post-training quantization without significant accuracy loss~\cite{krishnamoorthi2018quantizing_cnn}.
While traditional approaches apply QAT as a separate fine-tuning stage, recent works explore unified paradigms.
Quantization-Aware Knowledge Distillation~(QKD) coordinates these objectives to preserve accuracy in low-bitwidth models ~\cite{kim2019qkd}, and collaborative multi-teacher frameworks have been proposed to further guide the learning of ultra-low bit-width networks~\cite{pham2023multi_teacher}. 
Building on these concepts, we fuse QAT with Knowledge Distillation from the first epoch of Nano-U training, a design choice detailed in Section~\ref{sec:optimization}.

\subsection{TinyML Hardware and Inference Frameworks}
\label{sec:tinyml}

TinyML refers to the deployment of machine learning models on microcontrollers and other severely resource-constrained devices, typically operating within milliwatt power budgets and kilobyte memory footprints~\cite{warden2019tinyml}.
While purpose-built MCUs with dedicated accelerators, such as GAP9~\cite{rossi2021gap9} and Google Coral Edge TPU~\cite{cass2019coral}, achieve high efficiency with specialized integer arithmetic units, they often require proprietary software toolchains and higher costs than ubiquitous general-purpose MCUs.
In this work we target commodity off-the-shelf components that lack dedicated ML hardware but can be sourced for as little as \$10, testing the limits of robust perception on non-specialized platforms.

The efficiency of the inference framework used to execute ML models on an MCU is critical, as its memory overhead and execution strategy directly dictate system performance and deployment feasibility.
The most widely adopted one is TensorFlow Lite Micro~(TFLM)~\cite{david2021tflm}, an interpreter-based port of TensorFlow Lite designed for bare-metal embedded targets.
While this is flexible and hardware-agnostic, it carries non-negligible overhead
since the interpreter loop, its operator registry, and the internal bookkeeping compete directly with the tensor arena for the same memory budget. 

To eliminate such overhead, we adopt MicroFlow~\cite{carnelos2025microflow}, a compiler-based inference engine that resolves model topology at compile time via Rust procedural macros. 
In particular, in this work we extend MicroFlow with MaxPool2D and nearest-neighbor upsampling operators, as they are required by Nano-U's architecture.

\section{SYSTEM IMPLEMENTATION}
\label{sec:system}

\subsection{Nano-U Network Design}
\label{sec:design}

Nano-U is a binary segmentation network inspired by the popular U-Net architecture~\cite{ronneberger2015unet}. It is designed for devices with highly constrained memory like the ESP32-S3 MCU, using its 320~KB internal RAM budget as the primary limiting factor.
Its topology consists of a seven-stage encoder-decoder pipeline, summarized in Table~\ref{tab:topology}.

The network uses as input $60{\times}80{\times}3$ color images, the largest resolution that allows the encoder-decoder stack to fit within the available RAM.
Spatial dimensions and channel widths are designed so that the largest intermediate activation tensor is $60{\times}80{\times}4$~(19.2~KB in INT8), ensuring that the total tensor arena remains within the hardware bounds.
For the decoding stages, we use nearest-neighbor upsampling, as it allows exact integer memory copies with no multiply-accumulate operations, making it both hardware-efficient and numerically stable under INT8 quantization.

\begin{table}
\centering
\caption{Nano-U Encoder-Decoder Topology}
\label{tab:topology}
\begin{tabular}{lcccc}
\hline
\textbf{Stage} & \textbf{Input} & \textbf{Output} & 
\textbf{Ch.} & \textbf{Op.} \\
\hline
Encoder 1 & $60\times80$ & $30\times40$ & $3{\to}4$  & MaxPool $2\times2$ \\
Encoder 2 & $30\times40$ & $15\times20$ & $4{\to}8$  & MaxPool $2\times2$ \\
Encoder 3 & $15\times20$ & $5\times10$  & $8{\to}16$ & MaxPool $3\times2$ \\
Bottleneck & $5\times10$ & $5\times10$ & $16{\to}16$ & DW-Sep Conv \\
Decoder 1 & $5\times10$  & $15\times20$ & $16{\to}8$ & NN $3\times2$ \\
Decoder 2 & $15\times20$ & $30\times40$ & $8{\to}4$  & NN $2\times2$ \\
Decoder 3 & $30\times40$ & $60\times80$ & $4{\to}1$  & NN $2\times2$ \\
\hline
\end{tabular}
\end{table}

In contrast to U-Net, skip connections are omitted as retaining encoder feature maps in RAM until the decoder consumes them would easily exhaust the tight memory budget. 
This also fits MicroFlow's static allocation model, as no buffer must outlive the next layer, limiting peak memory use to a single active tensor buffer at any point.

Nano-U uses only depthwise separable convolutions, which reduce both parameter count and Multiply-Accumulate~(MAC) operations by a factor of $1/C_\text{out} + 1/K^2$ relative to standard convolutions, where $C_\text{out}$ is the number of output channels and $K$ is the kernel size. 
All layers apply this factorization with $K{=}3$.
The output is passed through a sigmoid function that produces per-pixel probabilities in the $[0,1]$ range. 
A threshold of $0.5$ is applied to the sigmoid output to produce the binary mask.
The original FP32 model contains 4,688 parameters, however at TFLite export time, batch normalization statistics are folded into the preceding convolutional weights and adjacent layers are fused~\cite{carnelos2025microflow}, reducing the stored parameter count to 3{,}357, without any loss of representational capacity.

\subsection{Optimization Pipeline}
\label{sec:optimization}

\begin{figure}
    \centering
    \includegraphics[width=\columnwidth]{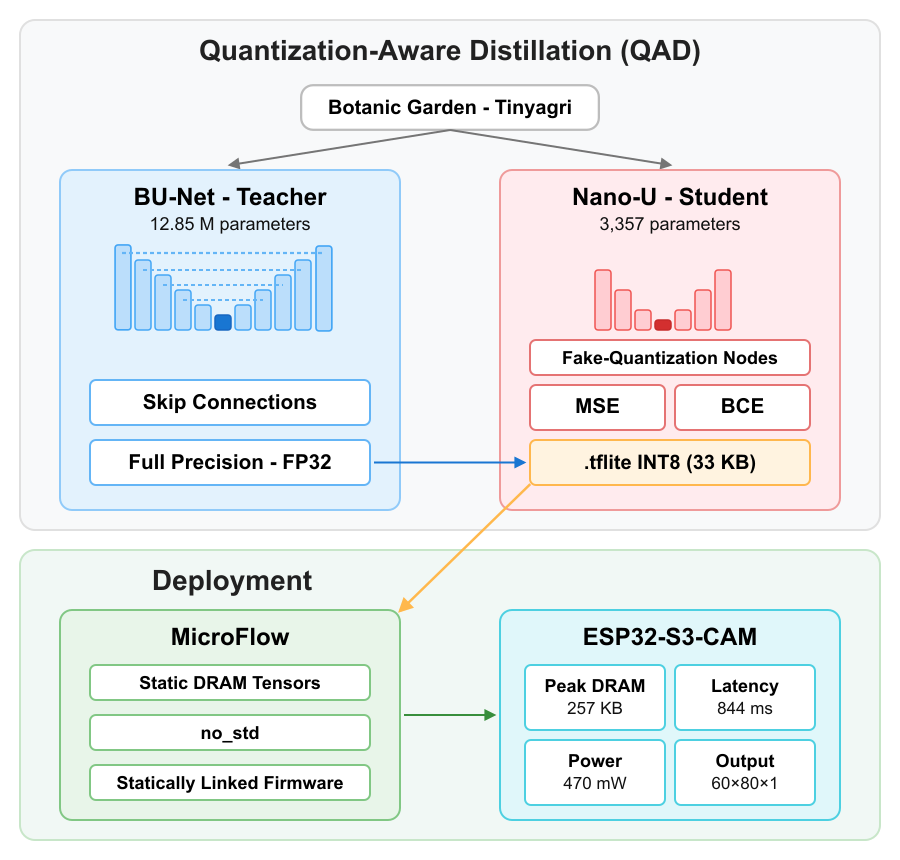}
      \caption{We train Nano-U via Quantization-Aware Distillation (QAD) where knowledge is distilled from a full-scale teacher while fake-quantization nodes simulate INT8 precision. The resulting model is compiled in Rust via MicroFlow for efficient execution on microcontrollers.}
      \label{fig:pipeline}
\end{figure}

Training Nano-U presents two concurrent challenges: the extremely small parameter budget limits representational capacity when trained on hard labels alone, and the INT8 deployment target introduces accuracy degradation if quantization is not accounted for during the training.
While recent frameworks explore quantization-distillation co-training~\cite{kim2019qkd, pham2023multi_teacher}, our pipeline adapts Quantization-Aware Distillation~(QAD)~\cite{polino2018dist_and_quant, xin2026nvidia_qad} to an INT8 binary segmentation network under an extreme parameter budget (Fig.~\ref{fig:pipeline}). 

Following Polino et al.~\cite{polino2018dist_and_quant}, we adopt a single-pass fused objective that simultaneously optimizes two losses from the first epoch: the hard cross-entropy loss against ground-truth labels, and the soft knowledge distillation loss from the teacher.
At the same time, fake-quantization nodes are injected into the training graph to simulate 8-bit integer arithmetic during the forward pass.
Gradient updates, however, are computed in full precision via the straight-through estimator, allowing the optimizer to minimize the distillation loss and the quantization error within a single unified training loop.

For the distillation step we use as teacher BU-Net (Big U-Net), a standard U-Net model with five max-pooling stages and channel widths $[64, 128, 256, 512, 1024]$, totalling 12.85~M parameters.
Since Nano-U performs binary segmentation with a single output logit, temperature scaling is applied in sigmoid space rather than the traditional softmax space:
\begin{equation}
    \tilde{p} = \sigma\!\left(\frac{z}{T}\right) = 
    \frac{1}{1 + \exp(-z/T)}
    \label{eq:sigmoid_temp}
\end{equation}
where $z$ is the pre-sigmoid logit and $T > 0$ controls the softness of the output. 
The distillation loss $\mathcal{L}_\text{KD}$ is the mean squared error between the temperature-scaled sigmoid outputs of teacher and student:
\begin{equation}
    \mathcal{L}_\text{KD} = \left(\sigma\!\left(\frac{z_t}{T}\right) - 
    \sigma\!\left(\frac{z_s}{T}\right)\right)^2
    \label{eq:kd_loss}
\end{equation}
where $z_t$ and $z_s$ are the teacher and student logits respectively.
The total training loss is:
\begin{equation}
    \mathcal{L} = \alpha \cdot T^2 \cdot \mathcal{L}_\text{KD} 
    + (1 - \alpha) \cdot \mathcal{L}_\text{CE}
    \label{eq:total_loss}
\end{equation}
where $\mathcal{L}_\text{CE}$ is binary cross-entropy against the hard ground-truth mask and $\alpha \in [0,1]$ controls the relative contribution of each objective. 
Temperature $T$ flattens the output distribution, exposing gradient signal on ambiguous pixels near crop-soil boundaries that a hard label would suppress. 
The $T^2$ factor compensates for the reduction in gradient magnitude induced by temperature scaling~\cite{hinton2015kd}.

Some quantization-aware distillation schemes drop the hard-label term~\cite{zhao2024ssqakd} for classification models with a full-precision self-teacher.
In our setting instead, with a distinct teacher and MSE-on-sigmoid distillation, retaining the CE term raises Precision by roughly one point at a small Recall cost, a trade-off we favor for safety-critical traversability.
Simultaneously, to prevent the disproportionate accuracy loss of post-training INT8 quantization on compact networks~\cite{krishnamoorthi2018quantizing_cnn}, 
QAT~\cite{jacob2018quantization_integer} is integrated from the first epoch. 
This joint objective is especially critical for structured spatial outputs, 
where quantization errors compound across the decoder~\cite{liu2019structured_kd}.

\subsection{Rust-based Deployment}
\label{sec:deployment}

After training Nano-U via QAD, the model is exported by calibrating INT8 scale factors and zero-points on the validation split, producing a 33~KB \textit{.tflite} file.
The model is compiled in Rust and deployed using MicroFlow, which is specifically designed for TinyML applications~\cite{carnelos2025microflow}.

A procedural macro generates a fully unrolled, statically-typed inference function with no operator registry, dynamic dispatch, or heap allocator, making all tensor buffers and quantization parameters compile-time constants~\cite{rustbook}.
Operating in a \textit{no\_std} Rust environment, the build runs directly on bare metal without any operating system~(OS) or runtime, activating only the peripherals strictly required for inference.
Unlike the standard TFLM ESP32 deployment, which requires FreeRTOS, non-volatile storage, and an event loop, our \textit{no\_std} build avoids these background tasks and unused subsystems, lowering energy per inference.

\section{EXPERIMENTAL SETUP}
\label{sec:exp}

\subsection{Target Hardware}
\label{sec:target}

All on-device experiments are conducted on a small \mbox{ESP32-S3-CAM}, a commodity module with an on-board camera~\cite{esp32}.
The processor features a dual-core Xtensa LX7 running at up to 240 MHz, equipped with a single-precision floating-point unit (FPU). 
However, because the latter accelerates only FP32 scalar operations, it does not provide hardware support for INT8 multiply-accumulate~(MAC) instructions required for quantized network inference\footnote{While the ESP32-S3 includes SIMD vector extensions for AI-acceleration, these primarily target FP32 operations via the vendor's proprietary toolchains, inaccessible from our bare-metal Rust \textit{no\_std} environment and orthogonal to INT8 inference.}.
Consequently, all integer operations of Nano-U execute on the general-purpose ALU, making INT8 quantization advantageous for both memory footprint and compute throughput, thereby reducing both latency and energy per operation.

The system provides 512 KB of internal static RAM, divided into approximately 320 KB of Data RAM (tensor arena, stack, static buffers) and 192 KB of Instruction RAM (executable code).
To establish a rigorous baseline, we constrain inference to a single core, leaving the secondary core available for concurrent system tasks.
Memory consumption is measured via \emph{stack painting}, a direct hardware measurement of peak RAM that captures all MicroFlow allocations and call overheads, independent of static estimates.

Finally, to isolate pure inference latency from peripheral I/O during our benchmarks, the onboard OV2640 camera is not used for live-capture.
Instead, we use the module's external 16~MB PSRAM to store the input images.

\subsection{Datasets and Evaluation Metrics}
\label{sec:dataset}

To evaluate Nano-U across various outdoor environmental conditions, we consider two different datasets.

First, we establish our baseline using the Botanic Garden dataset~\cite{liu2023botanicgarden}, a comprehensive benchmark collected in a 48,000~$\mathrm{m^2}$ unstructured outdoor environment. 
For our evaluation, we select a subset of 1,181 images, representing all 5 distinct sequences provided with segmentation annotations. 
Finally, to adapt the dataset to our binary architecture, we generate traversability masks by extracting only the \textit{path} class annotations from the original ground truth.

\begin{figure}
    \centering
    \includegraphics[width=\columnwidth]{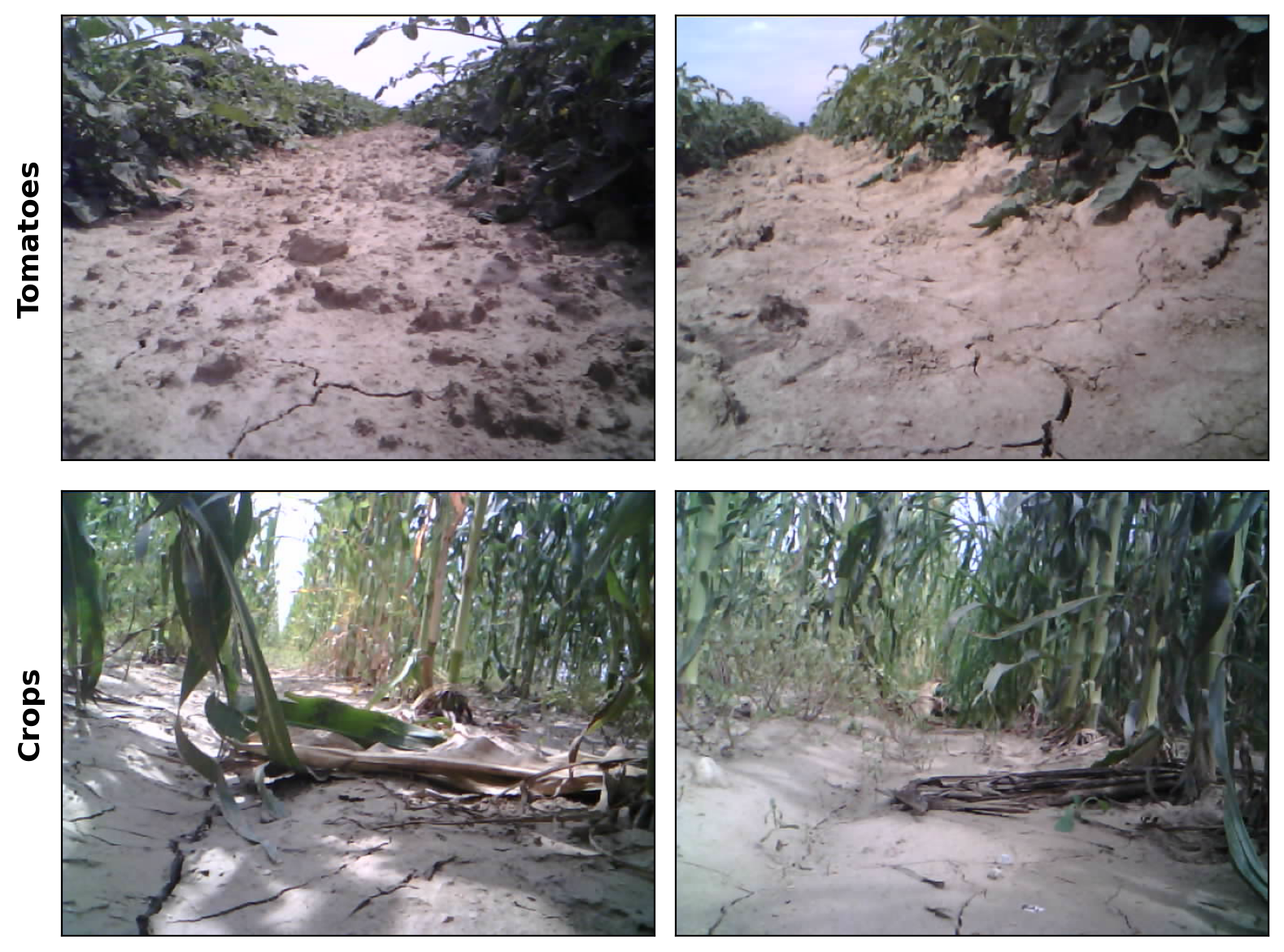}
    \caption{Representative frames from the TinyAgri dataset: the Tomatoes sequence (top) features a clear vanishing point and a more uniform row structure, while the Crops sequence (bottom) presents denser vegetation and greater leaf overlaps.}
    \label{fig:tinyagri_examples}
\end{figure}

Then, to evaluate the system under the same optical characteristics of our target hardware, we introduce TinyAgri, a custom terrain segmentation dataset collected via the on-board OV3660 camera of the ESP32-CAM board.
The dataset was collected using the SunFounder Galaxy RVR rover~\cite{sunfounder_galaxy_rvr}, shown in Fig.~\ref{fig:rover}, across two distinct agricultural environments.
Sampling from the camera at 1~FPS, we collected a total of 1,336 images across two sequences in tomato fields (Fig.~\ref{fig:tinyagri_examples} top), and 1,323 images in corn fields (Fig.~\ref{fig:tinyagri_examples} bottom).
TinyAgri introduces harder segmentation problems due to dense vegetation, greater leaf overlap, motion blur, and overall poorer image quality of the onboard camera.
We annotate the binary traversability labels using the SAM2 model~\cite{ravi2024sam2}.

To prevent data leakage in the datasets, we partition them by contiguous whole sequences rather than by random frame sampling, using a $70/20/10$ ratio for training, validation, and test sets, respectively. 
Although a few Botanic Garden routes partially overlap, their sequences were recorded on different days and times, so spatial leakage is negligible.
TinyAgri's sequences instead come from physically separate fields and are therefore spatially disjoint.

\begin{figure*}[t]
    \centering
    \includegraphics[width=\textwidth]{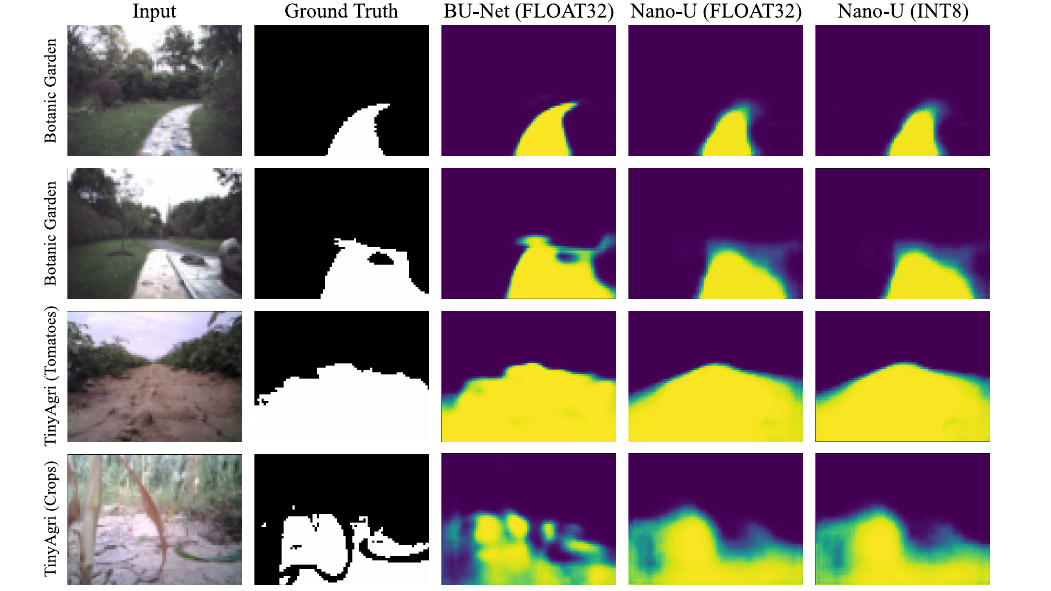}
    \caption{Qualitative comparison of segmentation masks. While Nano-U's extreme parameter constraints prevent it from capturing the fine-grained edge details of the SAM2 ground truth, it reliably identifies the core traversable terrain.}
    \label{fig:predictions}
\end{figure*}

The system performance of Nano-U is evaluated in terms of perception accuracy and hardware efficiency.
For perception accuracy, we use common metrics from the semantic segmentation literature~\cite{mo2022review_sota}: Mean Intersection over Union~(mIoU), Precision, Recall, and F1-score.
The first one, mIoU, measures the overlap between predicted and ground-truth binary masks averaged over both classes, penalizing false positives and negatives symmetrically, and providing a single scalar that is robust to class imbalance.
Precision and Recall respectively penalize over and under-segmentation of traversable regions.
These metrics are important for safety-critical navigation: a high-Precision, moderate-Recall model conservatively under-segments, causing the robot to decelerate at uncertain regions, whereas a low-Precision, high-Recall model might lead to untraversable paths.
Finally, F1-score provides their harmonic mean, summarizing the Precision-Recall trade-off in a single value.

Regarding hardware efficiency, the viability of Nano-U's deployment on the ESP32 board is quantified in terms of end-to-end inference latency~(ms), peak RAM utilization~(KB, measured via stack painting), and Power consumption~(mW, from the entire board power draw).

\subsection{Training Details}
All models are implemented in TensorFlow~2.20 and trained on a single NVIDIA GeForce RTX~5060 Laptop GPU.
The BU-Net teacher is trained from scratch on each dataset's training split for 200 epochs using a batch size of 32 and the AdamW optimizer, with an initial learning rate of $10^{-3}$ and a ReduceLROnPlateau schedule (factor $0.2$, patience $10$).
Nano-U is then trained via QAD for 200 epochs with a batch size of 16 and the same optimizer, with fake-quantization nodes injected from the first epoch. 
Its temperature and loss weight ($T{=}8.0$, $\alpha{=}0.3$) are selected by leakage-safe grouped $K$-fold cross-validation, ranking configurations by $F_{0.5}$ (the precision-weighted $F$-score, $\beta{=}0.5$) to weight Precision above Recall for the conservative, safety-critical operating point, with mIoU as a tiebreak and the test split held out.
During preprocessing, all images are scaled to the network's required $60\times80$ resolution, after which, to enhance generalization, we apply data augmentation with random horizontal flips and brightness and contrast jitter.
The final INT8 \textit{.tflite} model is exported by calibrating scale factors and zero-points on the validation split, leveraging the fake-quantization statistics accumulated during QAT, then compiled to a statically-linked Rust binary with MicroFlow.

For TinyAgri, Nano-U is retrained from scratch with the same pipeline; cross-validation instead selected a lower temperature and higher distillation weight ($T{=}4.0$, $\alpha{=}0.5$) consistent with its noisier camera frames and SAM2 labels, for which an over-softened teacher is less informative, so sharper soft targets and a stronger reliance on the distilled signal are favored.

\section{RESULTS}
\label{sec:results}

\subsection{Segmentation Accuracy in Botanic Garden}
\label{sec:botanic}

We first evaluate Nano-U on the Botanic Garden dataset (Fig.~\ref{fig:predictions}).
As reported in Tab.~\ref{tab:botanic}, the BU-Net teacher establishes a strong upper bound at 95.1\% mIoU, reflecting the high representational capacity of deeper networks, while the FP32 \mbox{Nano-U} student instead achieves 87.0\% mIoU, an excellent result due to the Knowledge Distillation approach, despite having $\sim4000 \times$ fewer parameters (12.85\,M vs.\ 3{,}357).
Crucially, the final INT8 model achieves 87.0\% mIoU, a negligible drop from its FP32 counterpart, confirming that the QAD pipeline successfully mitigates the typical accuracy degradation when mapping ultra-compact models to integer precision.
The high Precision with moderate Recall (0.807) reflects conservative under-segmentation, preferable for navigation: the robot avoids uncertain, collision-prone boundaries rather than committing to untraversable paths.

\begin{table}
\centering
\caption{Segmentation results on Botanic Garden}
\label{tab:botanic}
\begin{tabular}{lcccc}
\toprule
Model & mIoU & F1 & Precision & Recall \\
\midrule
BU-Net (Teacher)     & 0.951 & 0.957 & 0.962 & 0.952 \\
Nano-U (FP32)     & 0.870 & 0.876 & 0.960 & 0.806 \\
Nano-U (INT8, ESP32) & 0.870 & 0.877 & 0.960 & 0.807 \\
\bottomrule
\end{tabular}
\vspace{0.5cm}
\end{table}

\subsection{Segmentation Accuracy on TinyAgri}
\label{sec:tinyagri}

\begin{table}
\centering
\caption{Segmentation results on TinyAgri}
\label{tab:tinyagri}
\begin{tabular}{lcccc}
\toprule
Configuration & mIoU & F1 & Precision & Recall \\
\midrule
BU-Net (Teacher)     & 0.909 & 0.951 & 0.945 & 0.957 \\
Nano-U (FP32)     & 0.883 & 0.938 & 0.919 & 0.957 \\
Nano-U (INT8, ESP32) & 0.700 & 0.823 & 0.807 & 0.840 \\
\bottomrule
\end{tabular}
\end{table}

We then evaluate Nano-U on the custom TinyAgri dataset (Tab.~\ref{tab:tinyagri}), a harsher environment than Botanic Garden's structured paths, characterized by dense foliage and challenging agricultural terrain.

Again the FP32 Nano-U closely tracks BU-Net at 88.3\%~mIoU, confirming the effectiveness of Knowledge Distillation. 
On this harder dataset, the INT8 model shows a wider performance gap (70.0\% mIoU), although mostly concentrated in fine boundary details rather than traversable regions, retaining an F1 of 0.82 and a recall of 0.84. As Fig.~\ref{fig:predictions} shows, the model still delineates the core path structure, losing leaf-edge precision without missing the traversable area.
The quantized model therefore remains suitable for path planning on TinyAgri, with boundary fidelity the main focus for further improvement.

\subsection{Hardware Profiling}
\label{sec:profiling}

Tab.~\ref{tab:hardware} reports the hardware profiling results of \mbox{Nano-U} on the ESP32-S3-CAM and, for context, those of MobileNetV1-0.25, the reference classification model in MicroFlow~\cite{carnelos2025microflow}. 
Despite producing a dense $60{\times}80 $ binary segmentation mask, far more memory-intensive than binary classification, \mbox{Nano-U}'s inference latency is well below~1~second, with a peak RAM leaving approximately 60~KB free memory.

\begin{table}
\caption{Hardware Profiling on ESP32-S3-CAM}
\label{tab:hardware}
\centering
\setlength{\tabcolsep}{4pt}
\begin{tabular}{lcc}
\toprule
Metric & Nano-U & MobileNetV1-0.25\\
\midrule
Task                  & Binary segmentation & Binary classification\\
Parameters            & 3,357 & 210,708\\
Output                & $60{\times}80$ pixel mask & 1 scalar\\
Peak internal Data RAM & 257 KB & 170 KB\\
Inference latency     & 844 ms & 3,721 ms\\
\bottomrule
\end{tabular}
\end{table}

Finally, the deployment described in Sec.~\ref{sec:deployment} maximizes energy efficiency by keeping the wireless antenna, Bluetooth stack, and unused interfaces inactive throughout inference.
Board-level average power consumption during inference is 470~mW, compared to an idle baseline of 137~mW, representing an active inference overhead of just 333~mW, and an energy cost of 281~mJ~(333~mW $\times$ 844~ms).

\section{CONCLUSIONS}
\label{sec:conclusions}

This work demonstrates that real-time binary terrain segmentation for tiny mobile robots is achievable within the severe constraints of a commodity microcontroller.
We show that a relatively small network, Nano-U, can produce dense traversability masks on an ESP32-S3, consuming little memory and power.
The key enabler is the combination of quantization-aware distillation, which transfers structured spatial knowledge from a full-precision BU-Net teacher into an INT8 student with minimal loss on structured paths and usable accuracy on harder agricultural scenes. This is combined with MicroFlow's compile-time inference engine, which eliminates interpreter overhead and exposes the full memory and energy budget to the network.

The primary limitation of the system is inference latency: while 844~ms (i.e. 1.2~FPS) is sufficient for slow-moving ground platforms, it would not support faster vehicles or tasks requiring finer temporal resolution.
A second limitation is that the segmentation resolution ($60{\times}80$) is dictated by memory constraints rather than task requirements, and finer-grained masks could improve path planning precision.
Finally, the model was trained and evaluated on a single dataset at a time; generalization across vegetation types, lighting conditions, and soil textures remains to be validated.

Several directions remain open for future work.
The current evaluation isolates inference latency by loading images from PSRAM rather than the onboard OV2640 camera. Closing the full pipeline, live-capturing and processing images while controlling the robot's movement, would provide a complete picture of system-level feasibility for agricultural deployment.
Also, a preliminary analysis of the training process suggests Nano-U's encoder generalizes across terrain types while the decoder is domain-specific. This would motivate an efficient transfer strategy in which only the decoder is re-distilled on new data, reducing the computational overhead compared to a Nano-U trained from scratch.

\balance
\bibliographystyle{IEEEtran}
\bibliography{IEEEabrv, bibliography.bib}

\end{document}